\documentclass[letterpaper, 10 pt, conference]{ieeeconf}  % Comment this line out if you need a4paper

\IEEEoverridecommandlockouts                              % This command is only needed if 
\usepackage{amsmath} % assumes amsmath package installed
\usepackage{amssymb}  % assumes amsmath package installed
\usepackage[pdftex]{graphicx}
\usepackage{multirow}
\usepackage{amsfonts}
\usepackage{float}
\usepackage{pifont}
\usepackage{subcaption}
\usepackage{stfloats}
\usepackage{graphicx,adjustbox}

\usepackage{accents}
\usepackage{cite}
\usepackage{bm}

\usepackage[T1]{fontenc}
\usepackage[final]{changes}
\setdeletedmarkup{\textcolor{red}{\sout{#1}}}
\definechangesauthor[color=blue]{AM}
\definechangesauthor[color=brown]{VS}

\title{\LARGE \bf
HARMONI: Haptic-Guided Assistance for Unified Robotic Tele-Manipulation and Tele-Navigation
}

\author{Venkatesh Sripada$^{1}$, M Arshad Khan$^{2}$, Julia F{\"o}cker$^{2}$, Soran Parsa$^{3}$, Susmitha P, H Maior$^{4}$, Amir Ghalamzan E$^{1}$
\thanks{{$^{1}$University of Surrey, UK}, %{$^{2}$University of Lincoln, UK}, {$^{3}$University of Nottingham, UK}, {$^{4}$University of Huddersfield, UK}
{$^{2}$University of Lincoln, UK}, {$^{3}$University of Huddersfield, UK}, {$^{4}$University of Nottingham, UK}
}
}

\begin{document}

\maketitle
\thispagestyle{empty}
\pagestyle{empty}

%%%%%%%%%%%%%%%%%%%%%%%%%%%%%%%%%%%%%%%%%%%%%%%%%%%%%%%%%%%%%%%%%%%%%%%%%%%%%%%%
\begin{abstract}
Shared control, which combines human expertise with autonomous assistance, is critical for effective teleoperation in complex environments. While recent advances in haptic-guided teleoperation have shown promise, they are often limited to simplified tasks involving 6- or 7-DoF manipulators and rely on separate control strategies for navigation and manipulation. This increases both cognitive load and operational overhead. 
In this paper, we present a unified tele-mobile manipulation framework that leverages haptic-guided shared control. The system integrates a 9-DoF follower mobile manipulator and a 7-DoF leader robotic arm, enabling seamless transitions between tele-navigation and tele-manipulation through real-time haptic feedback. 
A user study with 20 participants under real-world conditions demonstrates that our framework significantly improves task accuracy and efficiency without increasing cognitive load. These findings highlight the potential of haptic-guided shared control for enhancing operator performance in demanding teleoperation scenarios.
% Shared control, combining human operators with autonomous systems, is vital for effective teleoperation in complex environments. Despite the advances in haptic-guided teleoperation of mobile manipulators, they are often applied to simplified problems with 6 or 7 DOF robotic manipulators {while using separate control methods for manipulation and navigation. This increases the cognitive load on the operator and complexity and operation overhead of the system}. This paper introduces a novel {unified tele-mobile-manipulation framework} which benefits from a haptic-guided shared control framework. The system integrates a 9-DoF follower mobile manipulator and a 7-DoF Leader Robotic Arm to provide haptic feedback, ensuring smooth transitions between tele-navigation and tele-manipulation.  We show our framework enhances operators' performance without increasing cognitive load. Validation with 20 participants in real-world conditions shows that haptic guidance improves task accuracy and efficiency for complex tele-mobile manipulation, highlighting the potential of this approach in challenging scenarios.
\end{abstract}

\section{Introduction} {M}{obile} manipulation, which integrates manipulators onto mobile robotic platforms, has become essential in domains such as agriculture and search-and-rescue. These systems are particularly valuable in hazardous or inaccessible environments where human presence is challenging. Despite significant advancements, achieving full autonomy in these complex scenarios remains a significant challenge~\cite{nielsen2007}.

Teleoperation of mobile manipulators provides a practical solution to bridge this autonomy gap. However, traditional teleoperation methods are often intricate and time-consuming. Shared control, a collaborative approach where human operators work alongside autonomous systems, offers a more efficient solution. In shared control setups, operators maintain control over the mobile robot and manipulator while autonomous systems assist during critical situations, such as avoiding collisions or navigating hazardous areas~\cite{luo2020ieee} {such as nuclear decommissioning sites}.

% reply to: Chnages made as replies to Associate Editor (AE) and Reviewer point 1 
However, critical factors such as haptic feedback via tactile sensors~\cite{dahiya2009tactile}—including radiation-robust technologies such as acoustic-based tactile sensing~\cite{Vishnu2024acoustic, parsons2024single}, and post-grasp optimality grasping~\cite{amir2017grasp}—minimum torque control~\cite{rahal2023haptic}, joint limit awareness~\cite{ghalamzan2017human}, grasping~\cite{parsa2020HapticGuided} informed by post-grasp collision~\cite{pardi2018choosing}, Leader-follower motion scaling~\cite{parsa2022chi} and optimal grasp planning~\cite{adjigble2019assisted, adjigble2023haptic} cannot be reliably inferred from indirect visual cues alone. This dependence on visual feedback imposes a significant cognitive burden on the operator, leading to fatigue and reduced task performance over time~\cite{marturi2016towards}, especially when teleoperating complex systems such as a 7-degree-of-freedom (DoF) manipulator.

 \begin{figure}[tb!]
	\centering
	\includegraphics[width = \linewidth]{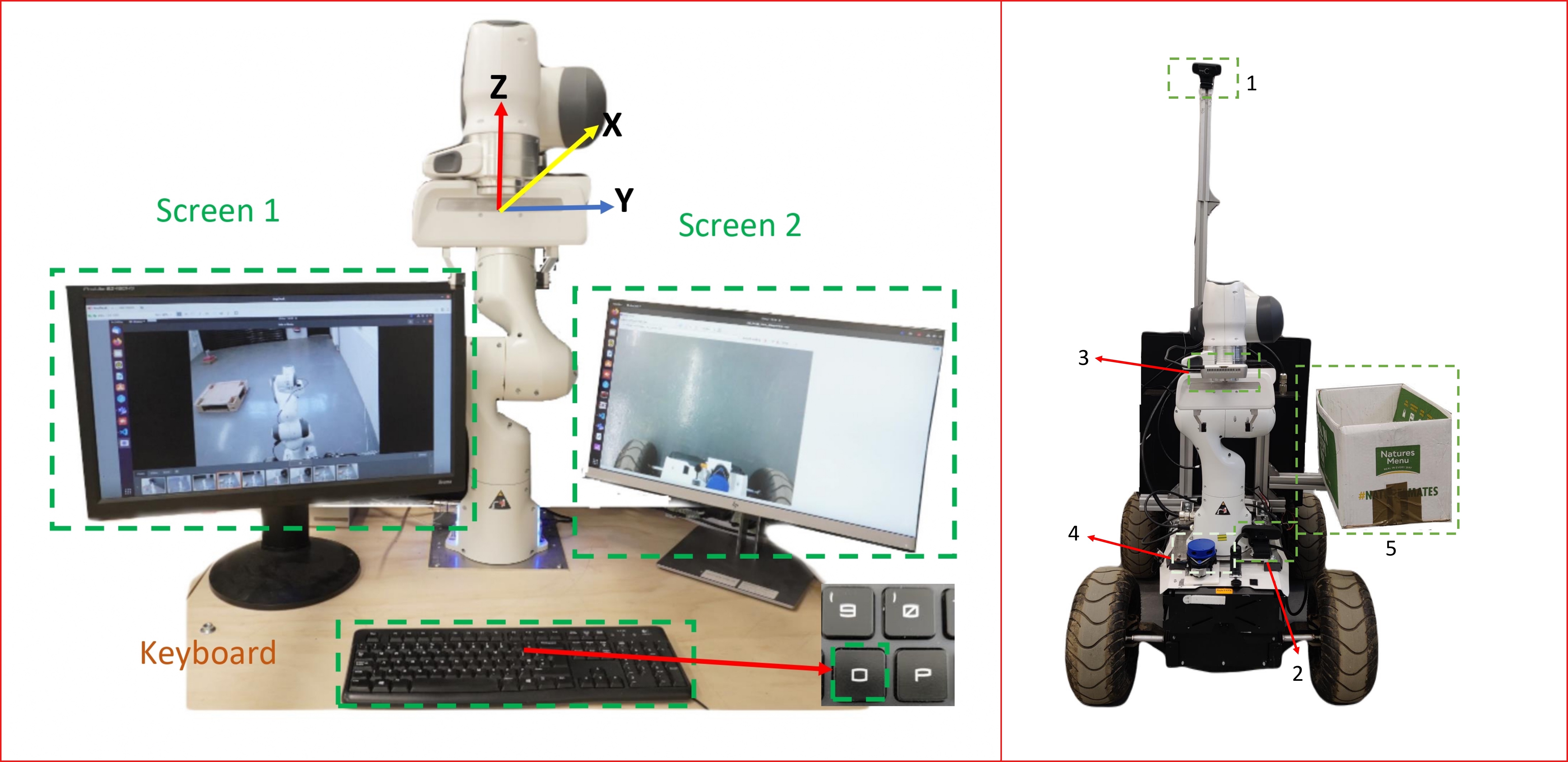}
	\caption{The operator controls the LRA (Panda robot on the left), with the FMR and FRA (Hunter 2.0 with Panda arm on the right) mirroring the LRA's movements. USB cameras $1$ and $2$ provide overhead and first-person views, while the Intel RealSense depth camera $3$ detects objects and offers an eye-in-hand view. The 2D lidar $4$ (YDLidar X4) assists with navigation, and the bin $5$ is used for autonomous object dropping after grasping.}
	\label{fig:la_fa}
\end{figure}

To address this challenge, we propose an integrated haptic guidance system with visual feedback and prove that it enhances control and reduces cognitive load during the teleoperation of mobile manipulators.
% Deleted the following line as it is the core of the paper. Addidng that we "hypothesised" it makes the paper look like it was bult on a vague hypothseis whereas thi paper was built on proven existing work as mentioned in the last paragraph of Related works

In this setup, the remote robot is referred to as the "follower-robot/arm," while the teleoperating device is the "leader-device/arm." The leader device, identical to the follower-arm, provides force feedback via impedance mode control, conveying critical information such as obstacles, joint limits, and non-holonomic constraints~\cite{rahal2019haptic, rahal2023haptic, ghalamzan2017human, parsa2020HapticGuided}.

While haptic guidance has proven effective in controlling various robotic systems, including manipulators, UAVs, and UGVs~\cite{ghalamzan2017human, parsa2020HapticGuided, rahal2023haptic, abi2019haptic, selvaggio2019haptic, varga2020shared, mohammadi2016cooperative}, its application in controlling both a UGV and a manipulator carried by the UGV is not well-explored.

This work introduces a novel shared control system that seamlessly manages both a Follower Mobile Robot (\textbf{FMR}) and a Follower Robotic \replaced[id=VS]{Arm}{Manipulator} (\textbf{FRA}) using a Leader Robotic Arm (\textbf{LRA}). The system provides haptic feedback to guide both UGV navigation and manipulator steering. We explore the complexities of integrating different types of haptic guidance (2-DoF for UGV navigation and 6-DoF for manipulator control) within a single system, significantly impacting how operators interpret and respond to these cues.

Our framework centres on guiding a mobile robot to an optimal path for stable object grasping by the FRA. We reduce equipment costs and streamline control transitions by utilizing a single LRA to control both the FMR and FRA, ensuring continuous operator engagement.

The contributions of this paper include (i) the development of a novel teleoperation system (Fig.~\ref{fig:la_fa}) consisting of a Franka Emika arm as the Leader device, teleoperating both mobile robot navigation and manipulator control; (ii) a 2-DoF impedance-controlled haptic-guided navigation system and a haptic-guided tele-manipulation algorithm for intuitive object grasping. We propose a seamless transition between tele-navigation and tele-manipulation. We validate the effectiveness of our proposed system with a human subject study. 

\section{Related Work}
Teleoperating robots span various domains, utilising both conventional control techniques and modern machine learning algorithms. A key concept employed in our methodology is the use of virtual fixtures, which provide force cues to the operator, guiding their actions to enhance performance by preventing deviations from predefined paths~\cite{wrock2017automatic, selvaggio2016enhancing}.

Prior research has explored teleoperation in simulation environments, where mobile manipulators execute motion plans based on a fully known environment~\cite{al2020human, 6DOFHapticMaster, mobilerobot2022iros}. However, these studies often do not account for real-world scenarios where unknown obstacles can emerge, highlighting the need for more adaptable solutions.

Recent advances in haptic feedback have primarily focused on tele-navigation and tele-manipulation tasks. In tele-manipulation, haptic force cues have been applied to inform operators about real-time events such as collisions, singularities, and joint limitations. These cues have been effective in both single-arm~\cite{abi2016visual} and dual-arm settings~\cite{selvaggio2018haptic}, as well as in guiding predicted trajectories of follower arms~\cite{pedemonte2017visual}. Haptic guidance has also been used to enforce nonholonomic constraints in tasks like robot-assisted cutting~\cite{rahal2019haptic}.

For tele-navigation, bilateral shared control frameworks with haptic feedback have been applied in various contexts, including disaster management. These frameworks typically allow for online trajectory adjustments where the human operator’s inputs are dynamically influenced by autonomous algorithms~\cite{masone2018shared}. Despite these advancements, the scope of such tasks remains limited to environments with known variables and reaching predefined goals~\cite{lui2017first}.

Haptic force cues have proven particularly useful in tele-manipulation involving leader and follower robots with high degrees of freedom (DoF) or kinematic similarities. For example, Singh et al.\cite{singh2020haptic} demonstrated the effectiveness of a 7-DoF leader arm in tele-manipulating a 7-DoF follower robot. The combination of haptic and visual guidance has been shown to reduce cognitive load and enhance operator performance, particularly when using identical leader and follower robotic arms\cite{rahal2023haptic, ghalamzan2017human}. Human subject tests further validate these findings, illustrating the efficacy of haptic guidance in improving tele-manipulation performance~\cite{parsa2022chi}.

Our research builds on these insights by integrating haptic guidance into both navigation and manipulation tasks, using a single leader robot identical to the follower robotic \replaced[id=VS]{arm}{manipulator}. This approach contrasts with prior methodologies that typically employ separate devices for navigation and manipulation~\cite{Xing2022MobileManipTimeDelay}. To assess the effectiveness of our system, we conduct comprehensive human subject tests using a real robot, evaluating its robustness in environments with unseen obstacles and under conditions that simulate real-world complexities, including sensory distractions.

\section{Problem Description and Methodology}
\label{sec:methodology}
Haptic-guided systems typically utilize specialized devices with limited Degrees of Freedom (DoF) for either tele-navigation or tele-manipulation tasks~\cite{masone2018shared, rahal2023haptic}. However, integrating both tele-navigation and tele-manipulation using a single Leader Robotic Arm (LRA) remains a challenge due to variations in robot morphology and the difficulty of generating meaningful force feedback through joint sensing.

Mobile robots can be effectively controlled with a 3-DoF joystick for movement in the x, y, and $\theta$ axes, while robotic arms generally require a 6-DoF device for teleoperation. The use of separate devices for these tasks complicates the operation. To address this, we propose a novel solution using a single 7-DoF robotic arm to control both the mobile robot and the manipulator. Our approach uses impedance control on the LRA to generate haptic force cues in the x-y plane, ensuring the LRA remains within its reachable workspace while also allowing for 2-DoF task space motion and stiffness control.

Let $P_o: {x_o, y_o, z_o, O_o}$ denote the pose of the marker on the object, $P_{fmr}: {x_{fmr}, y_{fmr}, z_{fmr}, O_{fmr}}$ the pose of the FMR, $P_{fra}: {x_{fra}, y_{fra}, z_{fra}, O_{fra}}$ the pose of the FRA's end effector, and $P_{lra}: {x_{lra}, y_{lra}, z_{lra}, \omega_{lra}, \theta_{lra}, \gamma_{lra}}$ the pose of the LRA's end effector. The LRA and FMR start from their respective home poses $P^{i}{lra}$ and $P^{i}{fmr}$. All poses $P \in \mathbb{R}^{6}$, where $x, y, z$ are positions, and $O$ represents the orientation vector.

\begin{figure*}[tb]
     \begin{subfigure}[tb]{1\textwidth}
         \centering   \includegraphics[width = .8\linewidth]{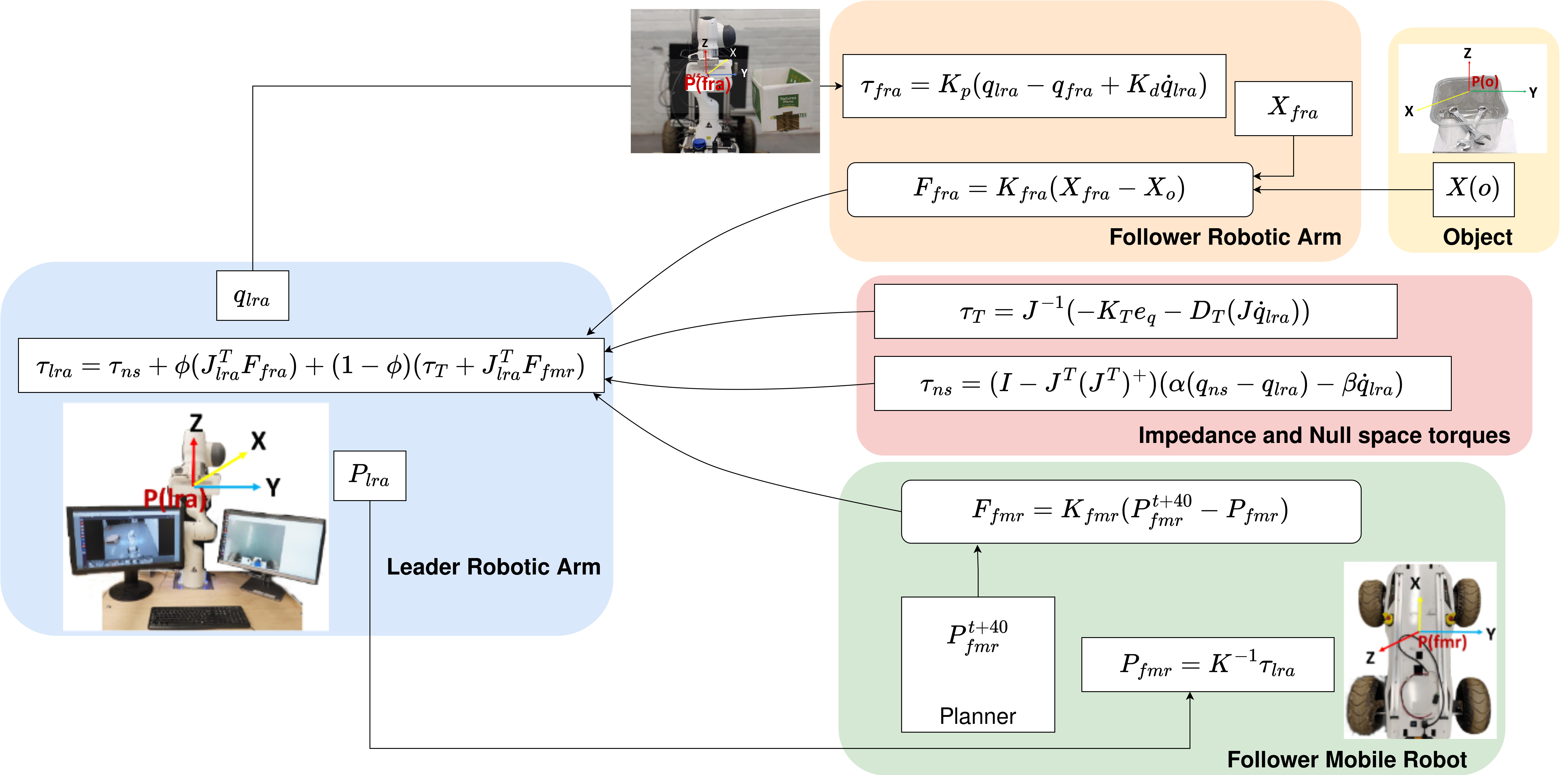}
         \caption{}
         \end{subfigure}
          \begin{subfigure}[tb]{1\textwidth}
         \centering
         \includegraphics[width = .78\linewidth]{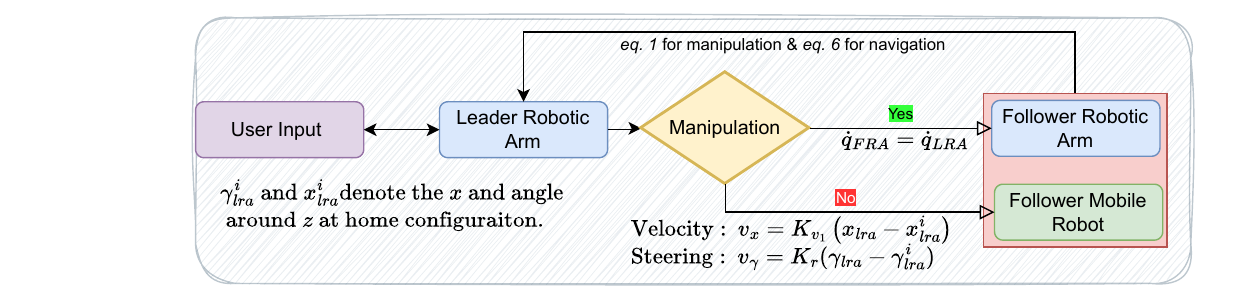}  
         \caption{}
         \end{subfigure}
	\caption{System architecture: (a) The Leader Robotic Arm (LRA) receives force input from either the Follower Robotic Arm (FRA) or the Mobile Robot (FMR), selected via the binary switch $\phi$. The FRA mirrors joint positions ($q_{lra}$), while the FMR’s velocity is derived from the LRA’s end-effector pose ($P_{lra}$). Impedance and null-space torques guide the LRA, with $X_{fra}, K_{fra}$ and $P_{fmr}, K_{fmr}$ denoting control targets and gains. (b) Control flow for switching between tele-manipulation and tele-navigation.}
%\caption{Architecture: The Leader Robotic Arm (LRA) receives force input from either the FRA or the FMR, controlled by the binary operator $\phi$. The FRA replicates the LRA’s joint positions ($q_{lra}$), while the FMR’s velocity is determined by the LRA’s end effector pose ($P_{lra}$). Impedance and null space torques guide the LRA, with $X_{fra}$ and $K_{fra}$ representing the FRA’s end effector position and gain, and $P_{fmr}$ and $K_{fmr}$ representing the FMR’s pose and gain.}
	\label{fig:architecture}
    \vspace{-.3cm}
\end{figure*}
%%%%%%%%%%%%%%%%%%%%%%%%%%%%%%%%%%%%%%%%%%%%%%%%%%%%%%%%%%%%%%%%%%%%%%%%%%%%%%%%
\emph{Architecture:} Our novel framework for controlling an integrated follower mobile robot equipped with a mirrored robotic arm is illustrated in Fig. \ref{fig:architecture}. To minimize time delays and ensure seamless teleoperation, the LRA and both FMR/FRA are connected to separate workstations via Ethernet cables. Both LRA and FRA are linked to the Franka Control Interface (FCI), maintaining a low-level bidirectional connection operating at 1 kHz. The robotic arms are programmed, allowing control through three modes: (i) joint torque commands (with gravity and joint friction compensation handled by FCI), (ii) joint positions and velocities, and (iii) Cartesian poses and velocities.

Communication between the leader and follower robots is facilitated using the Robot Operating System (ROS). The control script running on the LRA executes real-time control signals at 1 kHz, serving as the central point for execution and control. This setup ensures that the system operates efficiently with minimal latency.

At any given time, the torque \(\tau_{\text{lra}}\) applied to the LRA is determined by input from either the FRA or the FMR, depending on the task. This is controlled by a binary operator \(\phi\), where \(\phi = 0\) corresponds to FMR operation, and \(\phi = 1\) corresponds to FRA operation, as described in equation \ref{eq:tau_lra}:
\begin{equation} \label{eq:tau_lra}
    \tau_{\text{lra}} = \tau_{\text{ns}} + \phi \cdot (J_{\text{lra}}^{T} F_{\text{fra}}) + (1-\phi) \cdot (\tau_{T} + J_{\text{lra}}^{T} F_{\text{fmr}})
\end{equation}
Here, \(\tau_{\text{ns}}\) represents the null-space torques applied to avoid joint limits and singularities, while \(J_{\text{lra}}\) is the Jacobian of the LRA. The force feedback \(F_{\text{fra}}\) or \(F_{\text{fmr}}\) is generated based on the task—either tele-manipulation or tele-navigation.

For tele-manipulation, {having kinematic similarity increases operator interpretability \cite{singh2020haptic}. Hence, to avoid jerky motions during control} we employ a PD control approach to mirror the torques applied on the LRA at the FRA, as shown in equation \ref{eq:pd-mirror}:
\begin{equation} 
\label{eq:pd-mirror} 
    \tau_{\text{fra}} = K_{p} \cdot (q_{\text{lra}} - q_{\text{fra}}) + K_{d} \cdot \dot{q}_{\text{fra}}
\end{equation}
This ensures that the leader-follower arm movement is synchronised, with \(K_p\) and \(K_d\) being the proportional and derivative gains, \added[id=VS]{and $q$ and $\dot{q}$ being the joint positions and velocities}.

Null-space damping \(\tau_{\text{ns}}\) is applied to the LRA to prevent joint limit violations and avoid singularities, which in turn ensures the stability and smooth operation of the FRA. The formulation for null-space damping is given in equation \ref{eq:nullspace}:
\begin{equation}
    \label{eq:nullspace}
    \tau_{\text{ns}} = (I - J^{T} (J^{T})^{\dagger}) \cdot (\alpha \cdot (q_{\text{ns}} - q_{\text{lra}}) - \beta \cdot \dot{q}_{\text{lra}})
\end{equation}
where \(I \in \mathbb{R}^{7 \times 7}\) is the identity matrix, \(\alpha\) is the null-space stiffness, and \(\beta = 2 \sqrt{\alpha}\) is the null-space damping coefficient, which is tuned to achieve a damping ratio of 1.0. Here, \(q_{\text{ns}}\) is typically set equal to \(q_{\text{lra}}\), meaning the desired joint configuration is aligned with the current configuration. The matrices \(J\) and \(J^{\dagger}\) represent the Jacobian and its pseudo-inverse, respectively.

\noindent \emph{Tele-navigation:} The operator uses the LRA to control the FMR, with force cues provided for guidance. The LRA's translational and rotational movements along the $x_{lra}$ axis and $\gamma_{lra}$ rotation correspond to the FMR's motion. We employ virtual boundaries to prevent unintended motions and ensure the LRA returns to its home position when necessary. The linear and rotational velocities of the FMR are calculated using equations \ref{eq:7} and \ref{eq:8}.
\begin{equation} 
\label{eq:7} v_x = K_v\left( \frac{x_{lra} - x^{i}_{lra}}{VB{e} - VB_{i}}\right) \end{equation}
\begin{equation} \label{eq:8} 
v_{\gamma} = K_r (\gamma_{lra} - \gamma^i_{lra}) \end{equation}
where velocity scaling factor $K_v \in \mathbb{R}^1$ is {tuned and} set to $0.5$ when no obstacles are present and reduced to $0.2$ in the presence of obstacles. The angular velocity scaling factor $K_r \in \mathbb{R}^1$ is always set to $-1$. Virtual boundaries $VB$ are within the range $[VB_{i}, VB_{e}]$, where $VB_{i} = 0.05 m$ and $VB_{e} = 0.4 m$. $VB$s are distances along the x-axis from the home position beyond which the motion of the LRA no longer affects the FMR, and the LRA autonomously returns to its home position.

Specifically, once the LRA moves beyond $VB_{i}$, which corresponds to a distance of $5$ cm, the mobile platform is activated. This precaution ensures that the operator intentionally teleoperates the FMR, guarding against accidental displacement of the LRA. Conversely, if the LRA surpasses $VB_{e}$, equivalent to $40$ cm, it automatically returns to its home position. This behaviour is designed to prevent issues such as singularity and joint locking, which are part of Franka Panda's built-in safety features.

Haptic guidance for the leader-mobile platform control is provided based on the trajectory pose generated by ROS's DWA Planner. This guidance system ensures precise navigation and object grasping by the FRA, with force cues computed as shown in equation \ref{eq:9}:

\begin{equation} 
\label{eq:9} 
F_{fmr} = K_{fmr} (P^{t+40}_{fmr} - P_{fmr}) \end{equation}
where $P^{t+40}_{fmr} - P_{fmr} = [x^{t+40}_{fmr} - x_{fmr},0,0,0,0,\gamma^{t+40}_{fmr} - \gamma_{fmr}]$ and $K_{fmr} \in \mathbb{R}^{6\times6}$ is an emprically chosen gain.

% AE point 2, reviwer II, point 2$
{Haptic force cues, based on the target kinematic navigation trajectory, are generated using the DWA planner to account for both static and dynamic obstacles as well as changes in the map}

This architecture provides a robust solution for integrating tele-navigation and tele-manipulation into a unified control framework, ensuring efficient and intuitive operation of complex robotic systems.

\section{Teleoperation Setup} 
%%%%%%%%%%%%%
In this study, we introduce a 7-DoF Franka Emika Panda robotic manipulator to teleoperate both a Hunter 2.0 mobile robot and a follower Franka arm mounted on the Hunter 2.0. The LRA, positioned on a bench, controls both the Hunter 2.0 and the Franka arm (designated as FMR and FRA, respectively), as illustrated in Fig. \ref{fig:la_fa}. These follower robots are equipped with a YDlidar X4, an Intel RealSense depth camera, and two USB cameras that provide environmental information. The depth camera is mounted on the FRA's End-effector, serving as an eye-in-hand camera, while an Aruco marker is used for accurate object detection.

% Reply to reviewr II, point 1 
Our system operates in a static environment with a pre-generated map, 
{ allowing us to attribute observed changes in operator performance directly to our system design rather than to unpredictable external factors. We establish a baseline from which we can accurately compare the performance of our system when extended to more complex and dynamic scenarios in future studies. However, the underlying design of a unified haptic guided mobile manipulation framework remains valid in more complex environments.}

The Franka arm, a 7-DoF torque-controlled manipulator, features integrated torque sensors in each joint. The Hunter 2.0, a mobile robot by AgileX Robotics, has a payload capacity of 150 kg and a top speed of 10 km/hr, operating with Ackermann steering control.

The FMR continually updates Odometery and sensory data to account for dynamic changes, enabling the generation of global and local plans. {While Dijktra's algorithm was used to compute the global plan, DWA planner was used to generate local plans and haptic force cues as explained in Section \ref{sec:methodology}}. These plans are translated into distinguishable haptic force cues in Cartesian impedance space, which are conveyed to the operator via the LRA. This setup transforms the LRA into a haptic interface that relays both environmental and virtual information, enhancing teleoperation by providing intuitive feedback to the operator.
This study does not address stability concerns, as experiments were conducted under conditions ensuring stable operation.
 \begin{figure}[tb!]
	\centering
	\includegraphics[width = \linewidth]{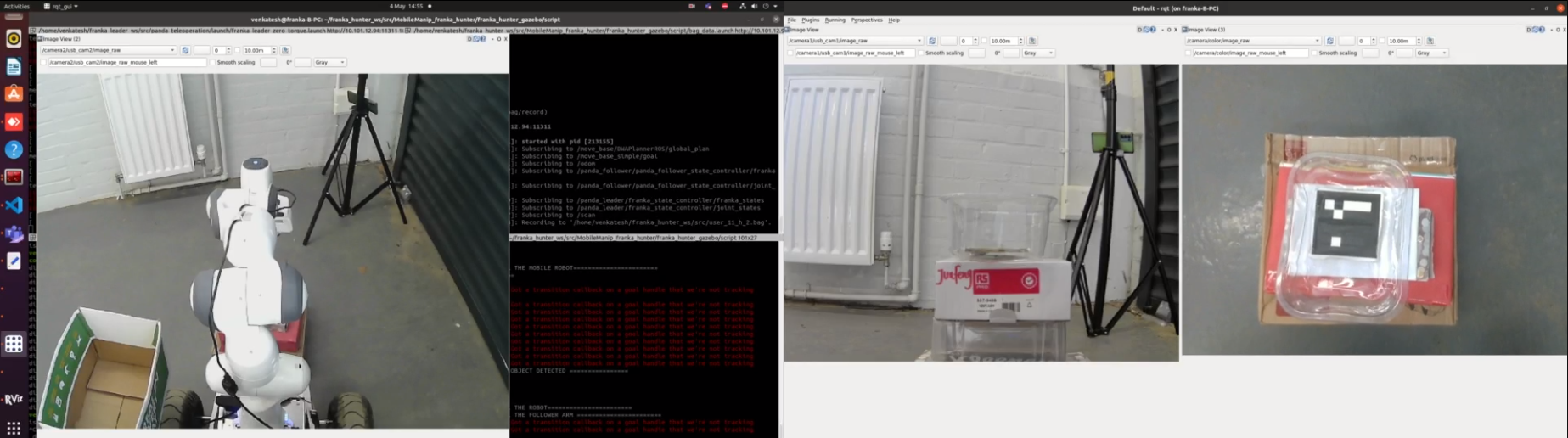}
	\vspace{-3pt}
	\caption{Operator's screens showing an overhead view (left), first-person view (centre), and eye-in-hand view (right) during teleoperation}
	\label{fig:la_screens}
    \vspace{-0.5cm}
\end{figure}

%  \begin{figure}
%      \centering
%      \begin{subfigure}[tb]{0.5\textwidth}
%          \centering
%          \includegraphics[width=\textwidth]{figures/screenshot_all_3_views.png}
%          \caption{Operators see 3 windows during teleoperation.}
%          \label{fig:la_screens}
%      \end{subfigure}
%      \hfill
%      \begin{subfigure}[tb]{0.3\textwidth}
%          \centering
%          \includegraphics[width=\textwidth]{figures/visual_distractor.png}
%          \caption{A red window appears on the screen for 1.5 secs.}
%          \label{fig:visual_distractor}
%      \end{subfigure}
%         \caption{Teleoperation Setup: (a) Operator's screens showing an overhead view (left), first-person view (centre), and eye-in-hand view (right) during teleoperation. (b) Visual distractor displayed during teleoperation.}
%         \label{fig:teleop_setup}
% \end{figure}

\begin{figure}[tb!]
	\centering
	\includegraphics[width = 0.92\linewidth]{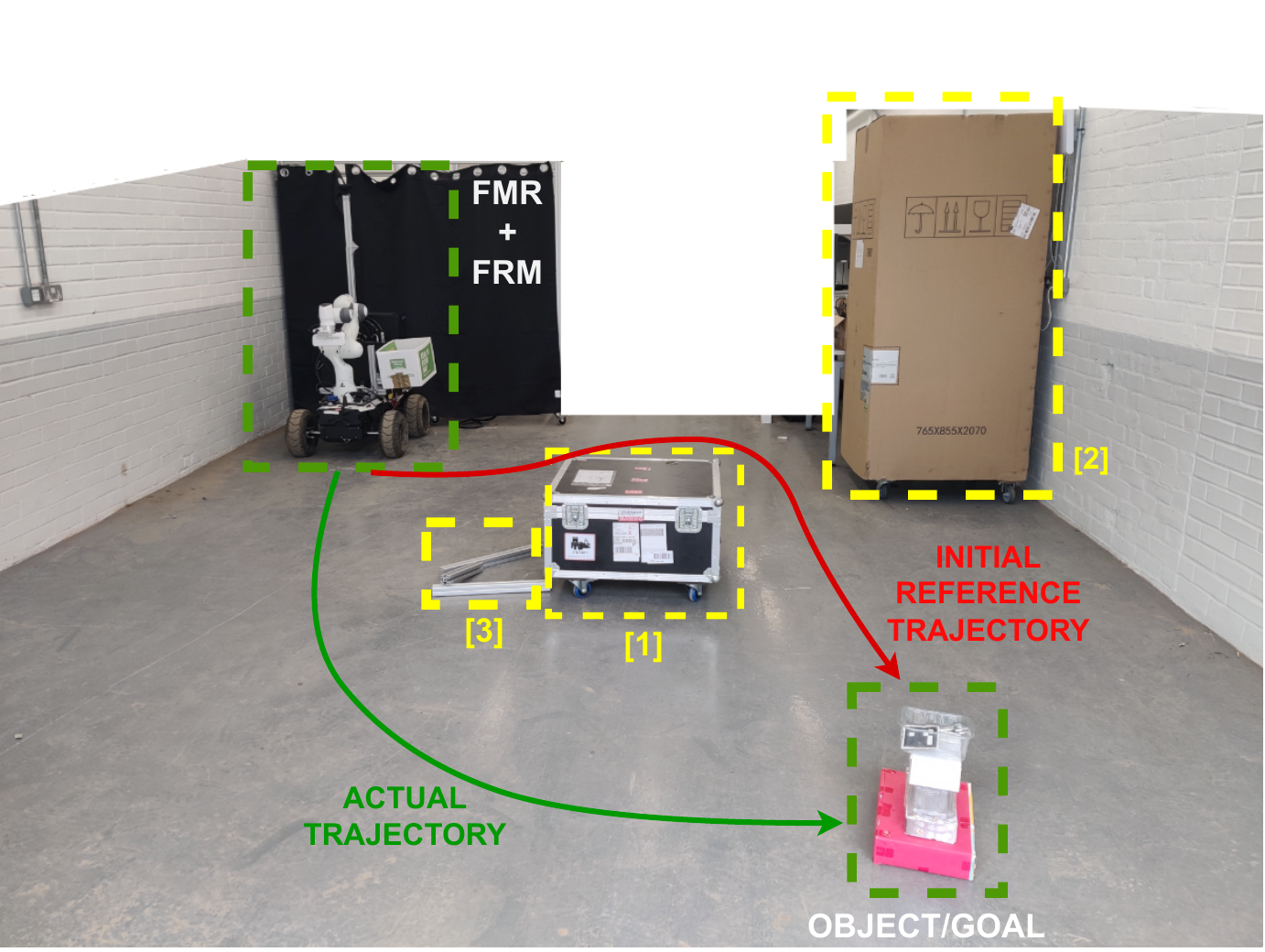}
	\vspace{-3pt}
	\caption{Map of the experiments: $1,2,3$ denote the \textit{known}, \textit{semi-known,} and \textit{unknown} obstacles respectively. The red and green arrows show the initial reference trajectory and the trajectory preferred by most participants}
	\label{fig:map}
    \vspace{-0.5cm}
\end{figure}

Visual feedback is provided to the operator via two screens on either side of the LRA, displaying an overhead view, first-person view, and eye-in-hand view, as shown in Fig. \ref{fig:la_screens}. The teleoperation process begins with controlling the FMR. Using a pre-known map (Fig.~\ref{fig:map}), the operator displaces the LRA, setting the FMR into motion, and a trajectory is planned towards the estimated object position, accounting only for known obstacles. The translational velocity of the FMR is capped at $0.5$ m/s to give the operator ample time to react, ensuring the safety of both the robot and its environment. 

% Reply to Reviewr II, point 4
\emph{Autonomous Switching:} As the FMR gets close to the object via operator's command, the eye-in-hand camera detects the Aruco marker on it. When the FMR enters the FRA's workspace and the object becomes \emph{graspable}, i.e. it is in an area from 20 cm to 45 cm in front of the FRA, the system autonomously switches from navigation mode to manipulation mode. This switch redirects the LRA's control to the FRA instead of the FMR, allowing the robotic arm control of the leader-follower. The operator is notified of this switch through both visual feedback and force cues, as the LRA joints momentarily stiffen, simulating a sense of locked joints. The LRA then returns to the home position and visual signal notify the operator that the switch is complete, i.e. $q_{lra}$ with $q_{fra}$. Once this alignment is confirmed, the joint impedance is reduced, allowing the operator to control the FRA. To minimize the risk of unintended switching due to perception errors, the Aruco marker ID on the target object and a force switching capability in navigation or manipulation mode was present, but hidden from the participants during the study.

This autonomous switching is advantageous as it offloads the task of determining object graspability from the operator, allowing them to focus on aligning the robot with the object, even if the optimal grasping position is not fully intuitive.

\paragraph*{Post-Grasp Manipulation} After grasping the object, the operator presses the keystroke $[o]$, initiating autonomous object dropping. The operator is again informed through haptic force cues, with the LRA's joints becoming temporarily locked due to maximised impedance. The FRA then autonomously plans and executes a trajectory to a bin at a fixed position on the mobile robot $P_{b}: {X_{b}, O_{b}}$. Once the object is dropped, the control reverts to the FMR. The LRA returns to the home position, and its joint impedance is reduced, enabling the operator to teleoperate the FMR towards the next object.

To validate the use of haptic guidance in teleoperating a mobile manipulator with an identical LRA, we conducted a human subject study with 20 participants. This study, approved by the University of Lincoln's ethics committee\footnote{Ethics Reference UoL2022 10345.}% the experiments involved participants signing informed consent forms covering privacy, safety, and their right to withdraw at any time}.

\begin{figure}
     \hfill
     \begin{subfigure}[tb]{0.5\textwidth}
        \centering
         \includegraphics[width=0.5\textwidth]{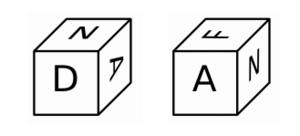}
         \caption{}
         \label{fig:cube_comp_test}
     \end{subfigure}
     \hfill
     \centering
     \begin{subfigure}[tb]{0.5\textwidth}
         \centering
         \includegraphics[width=0.5\textwidth]{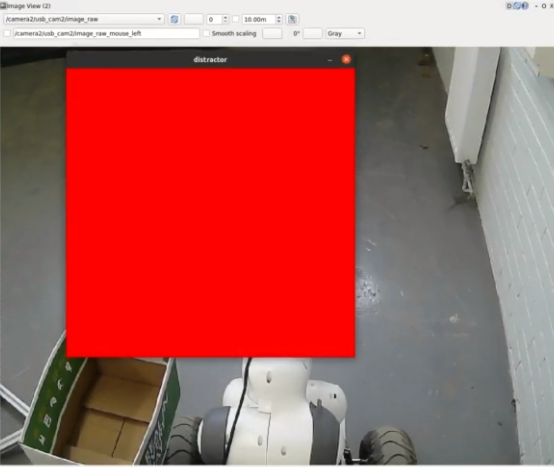}
         \caption{}
         \label{fig:visual_distractor}
     \end{subfigure}
        \caption{(a) Cube comparison test showing a dissimilar cubes example. Participants were asked whether cubes are similar(b); A visual distractor (a red window appears on the screen for 1.5 secs) appeared on screen during teleoperation.}
        % \label{fig:teleop_setup}
        \vspace{-.2cm}
\end{figure}

% Reply to reviewer I, point 2. 
\noindent We examined three parameters in our experiments:

\noindent \emph{Mental Rotation Ability:} Participants completed a six-minute cube comparison test (Fig. \ref{fig:cube_comp_test}) to assess their mental rotation ability. This test, consisting of 42 questions, helped determine any correlation between mental rotation skills and performance in mobile manipulation tasks. {The results were used for analysing the robustness of the system.}

%  \begin{figure}[tb!]
% 	\centering
% 	\includegraphics[width = 0.5\linewidth]{figures/cube_comp.png}
% 	\vspace{-3pt}
% 	\caption{Cube Comparison test example: Participants are asked whether cubes are similar. This shows a dissimilar cubes example. }
% 	\label{fig:cube_comp_test}
%     \vspace{-0.5cm}
% \end{figure}

\iffalse
\section{Experiments, Results and Discussion}
\noindent \emph{Audio Distraction:} To simulate real-world conditions, participants wore headgear playing a 40-second looped audio clip of a nuclear reactor startup. The volume was consistent across participants, adjusted only if discomfort was reported.

% Answering reviewer 2 who said that visual distractors were never explained
\noindent \emph{\added[id=VS]{Visual Distraction:}} \added[id=VS]{A red square window was overlaid on the operator’s display at random intervals for $1.5sec$ as shown in Fig. \ref{fig:visual_distractor}. This exposure was set based on pilot tests, which revealed that longer distractions disproportionately impaired operator performance and resulted in the robot colliding with previously unobserved obstacles.}
\fi
\section{Experiments, Results, and Discussion}

\noindent \textit{Audio Distraction:} To simulate realistic operating conditions, participants wore headgear playing a 40-second looped audio recording of a nuclear reactor startup. Volume was standardized across participants and adjusted only in cases of discomfort.

\noindent \textit{Visual Distraction:} A red square was intermittently overlaid on the operator’s screen for 1.5 seconds at random intervals (Fig.~\ref{fig:visual_distractor}). This duration was selected based on pilot tests, which showed that longer distractions significantly degraded performance and increased the likelihood of collisions with previously unobserved obstacles.

\iffalse
\noindent \emph{Heart Rate Variability (HRV):} HRV was measured using a Polar H10 monitor to assess the stress levels of participants during the experiments.

\noindent \emph{Experiment Design:}
Participants: Twenty participants (15 males, 5 females; mean age 27.6) with engineering backgrounds took part, including seven with prior robotics experience.

Three conditions were tested, with four trials per condition:

%

\noindent{\textbf{Condition 1 (Haptic Guidance)}:} Participants navigated the FMR to a goal, with haptic force cues provided via the LRA for guidance. Upon reaching the goal, the system autonomously switches to manipulation mode, enabling FRA control for the grasping task.

\noindent{\textbf{Condition 2 (No Haptic Guidance)}:} task execution without haptic force cues. Reference trajectories were computed and logged for later analysis.

\noindent{\textbf{Condition 3 (Visual Distractions)}:} This condition included visual distractions (Figure~\ref{fig:visual_distractor}), simulating a more challenging operational environment.

\noindent Conditions 1 and 2 were counterbalanced to prevent order effects. Participants were initially familiarized with the setup and performed a practice task with the object placed closer to avoid learning effects.

\vspace{.4cm}
\noindent \emph{Task Setup:} The FMR always started from the same pose, with a fixed object position (Fig. \ref{fig:map}). Operators controlled the LRA with no direct view of the FMR, relying solely on the screens provided.
\fi
\noindent \textit{Heart Rate Variability (HRV):} HRV was recorded using a Polar H10 sensor to assess participant stress levels throughout the experiment.

\noindent \textit{Experiment Design:}  
Twenty participants (15 male, 5 female; mean age 27.6), all with engineering backgrounds and seven with prior robotics experience, completed four trials under each of three conditions:

\begin{itemize}
  \item \textbf{Condition 1 (Haptic Guidance):} Participants navigated the FMR using haptic force cues provided via the LRA. Upon reaching the goal, the system automatically switched to manipulation mode, enabling FRA control for object grasping.
  \item \textbf{Condition 2 (No Haptic Guidance):} The same task was completed without haptic feedback. Reference trajectories were computed and logged for comparison.
  \item \textbf{Condition 3 (Visual Distractions):} Visual distractions (Fig.~\ref{fig:visual_distractor}) were introduced during task execution to simulate a more challenging environment.
\end{itemize}

Conditions 1 and 2 were counterbalanced to mitigate order effects. Participants received a familiarization trial using a simplified task to minimize learning bias.

\vspace{0.4cm}
\noindent \textit{Task Setup:}  
The FMR always began from the same initial pose, and the object was placed at a fixed location (Fig.~\ref{fig:map}). Participants operated the LRA using only visual feedback from provided screens, without direct line of sight to

\iffalse
We introduced \textit{known}, \textit{semi-known}, and \textit{unknown} obstacles to analyze the system's ability to adapt and generate accurate haptic force cues. Known obstacles were present in the prior map and detectable by lidar, semi-known obstacles were only detected when in range, and unknown obstacles were neither pre-mapped nor detectable. The initial reference trajectory was plausible but often suboptimal due to these obstacles. This setup allowed us to evaluate the effectiveness of the system in generating new trajectories and providing reliable haptic feedback, as well as understanding how participants processed this information alongside visual feedback.

Data from 160 trials (80 each for Condition 1 and Condition 2) were analyzed to evaluate force cue processing, ease of use, and stress response.

\noindent \emph{Force Cue Processing:} We assessed how participants responded to force cues by measuring the deviation between the actual and reference trajectories. Fig. \ref{fig:deviation} shows that during tele-manipulation, participants exhibited significantly lower deviation with haptic guidance along the y-axis $(p < 0.05, z = 2.012)$ (i.e. normal to the arc length of the planned trajectory). Haptic cues helped participants navigate more smoothly \added[id=VS]{(Fig. \ref{fig:meantrj})}, especially when manoeuvring around obstacles.
\fi

To evaluate the system’s adaptability and haptic feedback accuracy, we introduced three types of obstacles: \textit{known}, \textit{semi-known}, and \textit{unknown}. Known obstacles were included in the preloaded map and detected by lidar; semi-known obstacles became visible only within sensor range; and unknown obstacles were neither mapped nor detectable. The initial reference trajectory, while feasible, was often suboptimal due to these constraints, enabling assessment of the system’s capability to adapt and guide operators effectively.

We analyzed 160 trials (80 each from Conditions 1 and 2) to assess force cue interpretation, task efficiency, and stress response.

\begin{figure}[tb!] 
\centering 
\includegraphics[width=\linewidth, height=4.5cm]{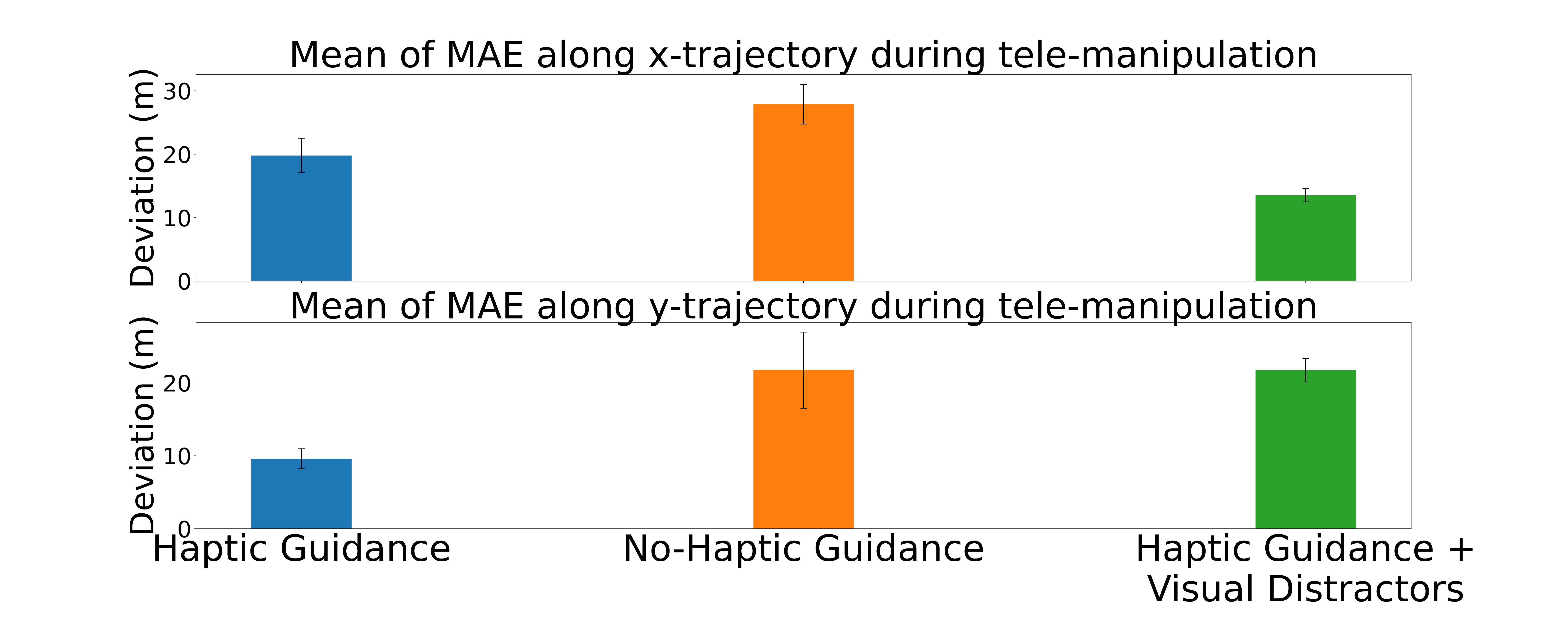}  
\caption{Mean deviation of actual trajectory from reference trajectory during tele-manipulation, \added[id=VS]{computed using mean absolute error}. Haptic guidance reduces deviation.} 
\label{fig:deviation} 
\end{figure}

\begin{figure}[tb!] 
\centering 
\includegraphics[width=1\linewidth, trim={6cm 0 4cm 0},clip]{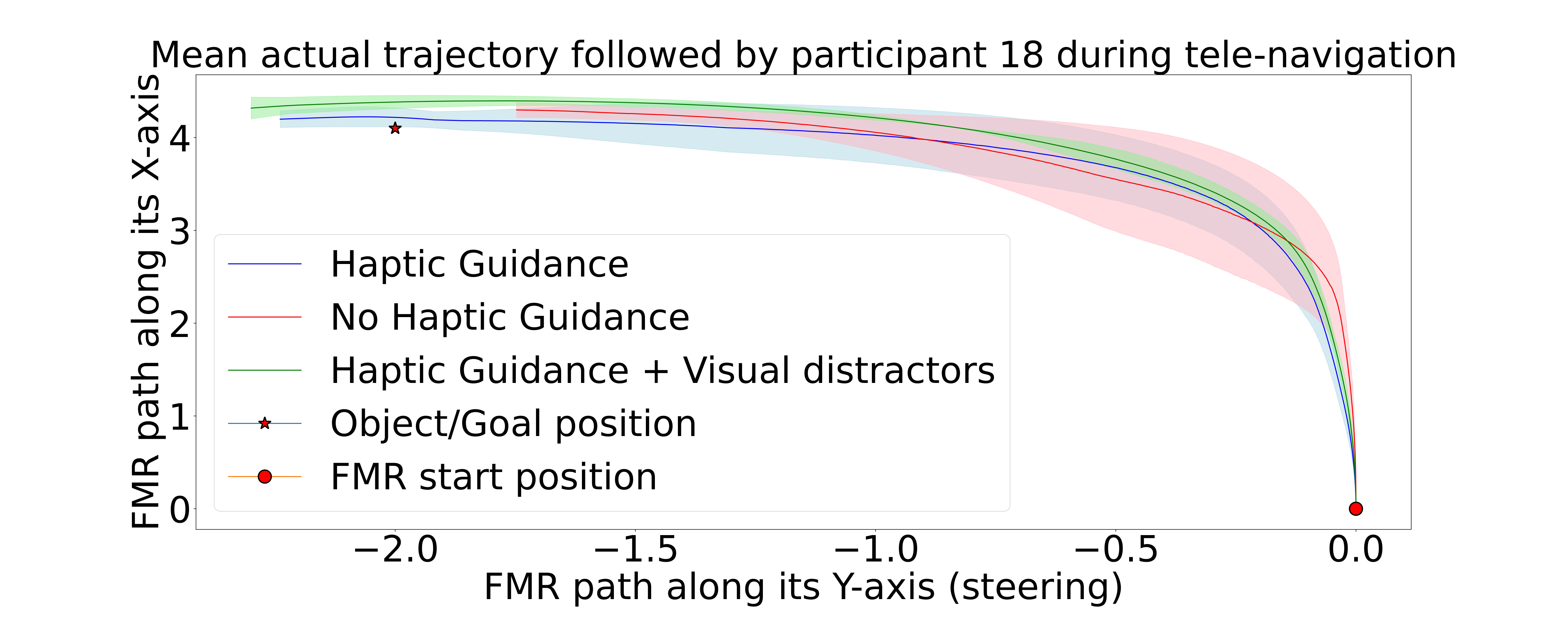}
\caption{Mean actual trajectory during tele-navigation shows smoother navigation with haptic guidance. \added[id=VS]{Shaded region shows variance}} 
\label{fig:meantrj}
\end{figure}

\noindent \textit{Force Cue Processing:}  
Participant responses to haptic guidance were quantified by measuring deviation from the reference trajectory. As shown in Fig.~\ref{fig:deviation}, during tele-manipulation, deviations along the y-axis were significantly reduced with haptic guidance ($p < 0.05, z = 2.012$)—i.e., orthogonal to the path arc. Haptic cues improved trajectory adherence and obstacle avoidance, particularly in cluttered scenes (Fig.~\ref{fig:meantrj}).

\begin{figure}[tb!]
	\centering
	\includegraphics[width = .9\linewidth, height=4.5cm]{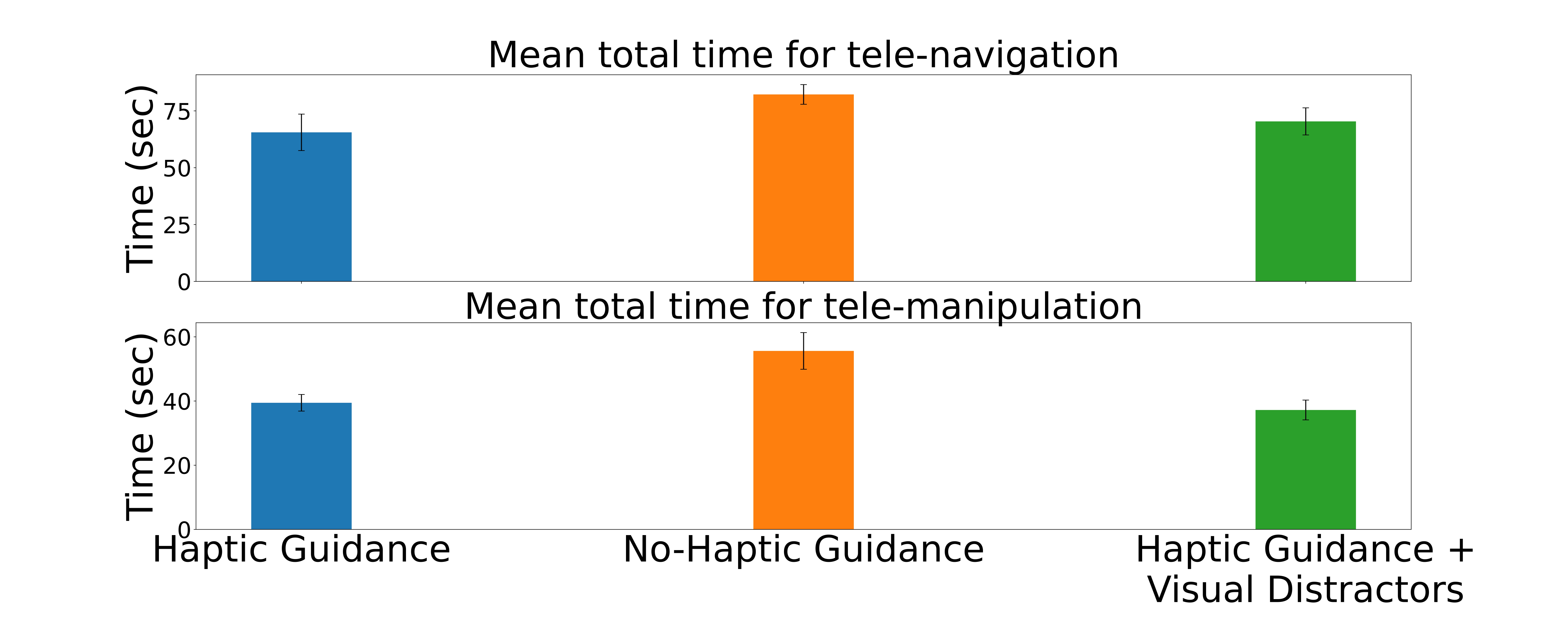}
	\vspace{-6pt}
	\caption{Mean total time for tele-navigation and tele-manipulation across users helps evaluate system usability. We observe that the use of haptic guidance is beneficial and significantly improves tele-manipulation. Error bars represent \added[id=VS]{Standard Error of Mean} (SEM).}
	\label{fig:teleop_3conds}
	\vspace{-6pt}
\end{figure}

\noindent \textit{Ease of Use:}  
Task completion time was used as a quantitative metric for usability, where shorter durations indicate reduced cognitive and operational effort. As shown in Fig.~\ref{fig:teleop_3conds}, haptic guidance significantly decreased task duration during tele-manipulation ($p < 0.05, z = 2.012$), though no significant difference was observed during tele-navigation ($p = 0.948, z = -0.065$). This suggests that haptic cues are particularly beneficial for complex manipulation tasks requiring higher precision.

\begin{figure}[tb!] 
\centering \includegraphics[width=.9\linewidth, height=4.5cm]{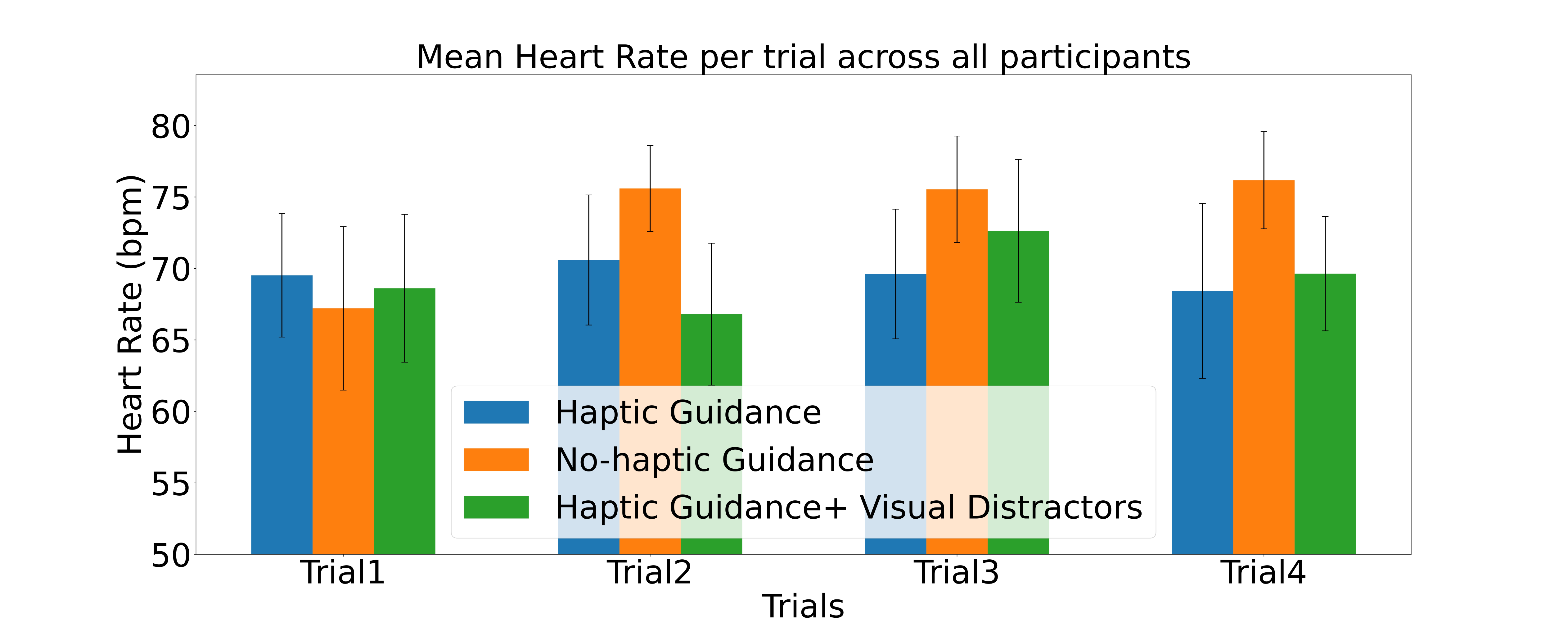}  
\caption{Mean heart rate across trials suggests reduced stress with haptic guidance.} 
\label{fir:heartrate} 
\end{figure}

\noindent \textit{Stress Response:}  
Heart rate (HR) was monitored as a physiological stress indicator~\cite{knight2001relaxing}. In Condition 1, 13 of the 20 participants exhibited lower HRs compared to Condition 2, with an average decrease of 1.05\%, indicating reduced stress levels when haptic cues were provided (Fig.~\ref{fir:heartrate}).

 \vspace{.3cm}

%\vspace{0.5cm}
\noindent \textit{Robustness (Condition 3 – Visual Distraction):}  
Under visual distractions, participants tended to select longer trajectories but navigated more smoothly. Analysis of 236 trials (Table~\ref{tab:deviation_p_values}) revealed reduced deviation from the reference trajectory with haptic guidance, highlighting its robustness even in visually perturbed environments.

Additionally, 11 of 18 participants showed lower HRs in Condition 3 than in Condition 2, further supporting the stress-buffering effect of haptic feedback. Results from the cube comparison test showed no significant correlation between mental rotation ability and task performance in Condition 3 ($r_s(18) = -0.055, p = 0.817$), suggesting that the system is accessible to users with diverse spatial reasoning skills.

 \vspace{.3cm}

 \vspace{.5cm}

\begin{table}[tb!]
    \centering
    \begin{tabular}{|l|c|c|} 
    \hline 
    \textbf{Comparison} & \textbf{Condition 1} & \textbf{Condition 2} \\
    \hline \hline
    \multicolumn{3}{|c|}{\textbf{Mean Deviation Along X-Axis}} \\
    \hline
    Condition 3 (Tele-navigation) & 0.687 & \textbf{p $<$ 0.05} \\
    Condition 3 (Tele-manipulation) & {p $<$ 0.05} & \textbf{p $<$ 0.001} \\
    \hline
    \multicolumn{3}{|c|}{\textbf{Mean Deviation Along Y-Axis}} \\
    \hline
    Condition 3 (Tele-navigation) & \textbf{0.107} & 0.155 \\
    Condition 3 (Tele-manipulation) & 0.831 & \textbf{p $<$ 0.05} \\
    \hline
    \end{tabular}
    
    \caption{P-values from the Wilcoxon signed-rank test for the mean deviation between actual and reference trajectories. Significant results indicate the advantage of haptic guidance (Condition 3) over no-haptic guidance (Condition 2), even under distraction.}
    \label{tab:deviation_p_values}
    \vspace{-1pt}
\end{table}

\vspace{-1.3cm}
\section{Conclusion}
In this study, we introduced a novel haptic-guided teleoperation framework that enables unified control of both a follower mobile robot (FMR) and a follower robotic arm (FRA) using a single leader robotic arm (LRA). A user study with 20 participants demonstrated that reliable haptic force cues enhance usability by improving trajectory adherence without increasing operator stress. The framework’s consistent performance across participants with diverse spatial reasoning abilities highlights its inclusivity and practical applicability.
Future work will extend the evaluation to include different types of leader devices and a larger participant cohort. We also plan to test the system in more complex, dynamic environments featuring moving obstacles and higher task variability. Additionally, comparative studies with dual-device control schemes—where separate leader interfaces are used for navigation and manipulation—will be explored to benchmark performance, although such comparisons lie beyond the core scope of this paper.

% In this study, we introduced a novel haptic-guided teleoperation framework that enables simultaneous control of both an FMR and FRA using a single LRA. Through a user study with 20 participants, we demonstrated that reliable force cues enhance usability, leading to improved trajectory adherence without increasing operator stress. The framework's effectiveness across participants with varying spatial reasoning abilities highlights its inclusivity and practicality. {Future work include evaluation of the algorithm's performance with various leader robots and a larger number participants. Preliminary tests indicate the technology's effectiveness, and we plan to conduct extensive human-subject experiments with more complex tasks and dynamic obstacles to confirm these findings. Future work will compare our method with other approaches that use separate leader devices for tele-navigation and tele-manipulation, although this comparison is beyond the core contribution of this paper.}
This research paves the way for broader adoption of haptic-guided shared control systems in complex and high-stress environments.

%%%%%%%%%%%%%%%%%%%%%%%%%%%%%%%%%%%%%%%%%%%%%%%%%%%%%%%%%%%%%%%%%%%%%%%%%%%%%%%%%%%%%%%%%%%%%%%%%%%%%%%%%%%%%%%%%%%%%%%%%%%%%%%%%%%%

\addtolength{\textheight}{-1cm}   % This command serves to balance the column lengths

%%%%%%%%%%%%%%%%%%%%%%%%%%%%%%%%%%%%%%%%%%%%%%%%%%%%%%%%%%%%%%%%%%%%%%%%%%%%%%%%

\bibliographystyle{IEEEtran}
\bibliography{ref}

\end{document}